\documentclass[letterpaper, 10 pt, conference]{ieeeconf}  

\IEEEoverridecommandlockouts                              

\usepackage{amssymb}  

\usepackage{cite}
\usepackage{amsmath,amssymb,amsfonts}
\usepackage{algorithmic}
\usepackage{graphicx}
\usepackage{textcomp}
\usepackage{xcolor}

\usepackage{makeidx}

\usepackage{pgfplots}
\pgfplotsset{compat=newest}
\usepackage{multirow}
\usepackage{tabularx}
\usepackage{threeparttable}
\usepackage{booktabs}

\usepackage[draft,inline,nomargin]{fixme}
\fxusetheme{color}

\usepackage{array}
\usepackage{float}
\usepackage{graphbox}
\usepackage{subcaption}
\usepackage{tikz,tikz-3dplot}
\usetikzlibrary{fit,arrows,arrows.meta,automata,backgrounds,calc,chains,%
decorations.markings,decorations.pathreplacing,decorations.pathmorphing,%
matrix,positioning,shapes,shapes.geometric,shapes.symbols,spy,trees,tikzmark}
\usepackage{balance}
\usepackage[binary-units=true,product-units=single,per-mode=symbol,range-units=single,range-phrase=\,--\,,detect-all]{siunitx}
\DeclareSIUnit\pixel{px}
\usepackage{bm}
\usepackage{acronym}
\usepackage{tablefootnote}
\usepackage{hyperref}

\usepackage{caption}
\let\labelindent\relax
\usepackage[inline]{enumitem}

\usepackage{adjustbox}
\newcolumntype{R}[2]{%
    >{\adjustbox{angle=#1,lap=\width-(#2)}\bgroup}%
    l%
    <{\egroup}%
}
\newcolumntype{L}[1]{>{\raggedright\let\newline\\\arraybackslash\hspace{0pt}}m{#1}}

\title{\LARGE \bf
DiffSSC: Semantic LiDAR Scan Completion using Denoising Diffusion Probabilistic Models
}

\author{Helin Cao and Sven Behnke
\thanks{
	This research has been supported by MBZIRC prize money. All authors are with the Autonomous Intelligent Systems group, Computer Science Institute VI – Intelligent Systems and Robotics – and the Center for Robotics and the Lamarr Institute for Machine Learning and Artificial Intelligence, University of Bonn, Germany; {\tt\small caoh@ais.uni-bonn.de}}%
}

\begin{document}

\maketitle
\thispagestyle{empty}
\pagestyle{empty}

\begin{abstract}
Perception systems play a crucial role in autonomous driving, incorporating multiple sensors and corresponding computer vision algorithms. 3D LiDAR sensors are widely used to capture sparse point clouds of the vehicle's surroundings. However, such systems struggle to perceive occluded areas and gaps in the scene due to the sparsity of these point clouds and their lack of semantics. To address these challenges, Semantic Scene Completion (SSC) jointly predicts unobserved geometry and semantics in the scene given raw LiDAR measurements, aiming for a more complete scene representation. Building on promising results of diffusion models in image generation and super-resolution tasks, we propose their extension to SSC by implementing the noising and denoising diffusion processes in the point and semantic spaces individually. To control the generation, we employ semantic LiDAR point clouds as conditional input and design local and global regularization losses to stabilize the denoising process. We evaluate our approach on autonomous driving datasets, and it achieves state-of-the-art performance for SSC, surpassing most existing methods.
\end{abstract}

\section{Introduction}
\label{sec:Introduction}
Perception systems collect low-level attributes of the surrounding environment, such as depth, temperature, and color, through various sensor technologies. These systems leverage machine learning algorithms to achieve high-level understanding, such as object detection and semantic segmentation. 3D LiDAR is widely used in self-driving cars to collect 3D point clouds. However, 3D LiDAR has inherent limitations, such as unobservable occluded regions, gaps between sweeps, non-uniform sampling, noise, and outliers, which present significant challenges for high-level scene understanding.

To provide dense and semantic scene representations for downstream decision-making and action systems, Semantic Scene Completion (SSC) has been proposed, aimed at jointly predicting missing points and semantics from raw LiDAR point clouds. Given its potential to significantly improve scene representation quality, this task has garnered significant attention in the robotics and computer vision communities. Understanding 3D surroundings is an inherent human ability, developed from observing a vast number of complete scenes in daily life. When humans observe a scene from a single view, they can leverage prior knowledge to infer unseen geometry and semantics. Drawing inspiration from this capability, 
the SSC model learns prior knowledge of scenes, $ \mathit{P}(\text{scene}) $, by estimating the complete scene from partial inputs during training. During inference, new partial inputs captured from the scene serve as the likelihood, $ \mathit{P}(\text{observation}|\text{scene}) $, and the model finally estimates a reasonable posterior result. Notably, the final estimation is not a unique answer but rather a sample from the posterior distribution, $ \mathit{P}(\text{scene}|\text{observation}) $. This aligns with intuition, since humans also infer plausible results from partial inputs, while the unobserved parts can have multiple possible completions.

\begin{figure}[t]
    \centering
    \begin{subfigure}[b]{0.23\textwidth}
        \includegraphics[height=0.10\textheight, width=\textwidth]{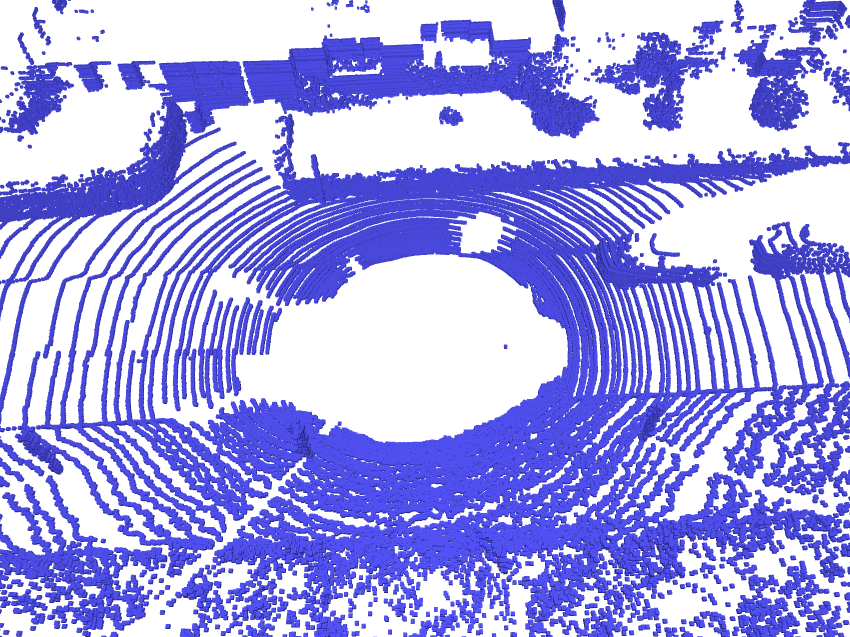}
        \caption{Sparse LiDAR Input}
        \label{fig:input}
    \end{subfigure}
    \hfill
    \begin{subfigure}[b]{0.23\textwidth}
        \includegraphics[height=0.10\textheight, width=\textwidth]{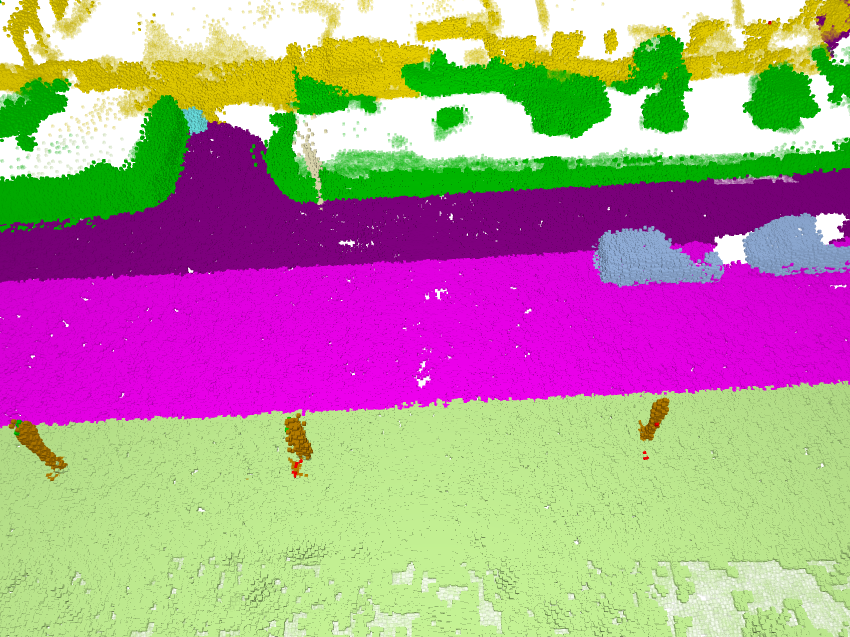}
        \caption{Dense Semantic Estimation}
        \label{fig:output}
    \end{subfigure}
    \caption{DiffSSC estimates unseen points with semantics (b) from raw LiDAR point clouds (a). The unknown areas, as defined by ground truth, are visualized at 20$\%$ opacity in (b).}
    \label{fig:teaser}
    \vspace{-1.5em}
\end{figure}

However, most traditional SSC methods are limited to learning the prior distribution of data directly, i.e., training a model to estimate the target output directly from partial inputs. Another approach to learning prior distributions is to estimate residuals. Denoising Diffusion Probabilistic Models (DDPMs) gradually inject noise into the data in the forward diffusion process and employ a denoiser to learn how to remove these noise residuals. The denoiser iteratively predicts and removes noise, allowing the model to recover high-quality data from pure noise. This mechanism effectively learns the prior distribution of the data, which has the potential to be applied in SSC tasks.


In this work, we propose DiffSSC, a novel SSC approach leveraging DDPMs. As shown in Fig.~\ref{fig:teaser}, our method jointly estimates missing geometry and semantics from a scene using raw sparse LiDAR point clouds. During training, the model learns the prior distribution by predicting residuals at different noise intensity levels. These multi-level noisy data are generated from ground truth using data augmentation. In the inference stage, the sparse semantic logits serve as conditional input, and the model generates a dense semantic scene from pure Gaussian noise through a multi-step Markov process. We model both the point and semantic spaces, designing the forward diffusion and reverse denoising processes to enable the model to learn the scene prior to the semantic point cloud representation. In summary, our key contributions are: 
\begin{itemize} 
	\item We utilize DDPMs for the SSC task, introducing a residual-learning mechanism compared to traditional approaches that directly estimate the complete scene from partial input. 
	\item We jointly model the noise injection process in both the spatial and semantic domains and design corresponding local and global regularization losses to enhance generation quality.
	\item Our approach operates directly on the point cloud, avoiding quantization errors and reducing memory usage, while making it a more efficient method for LiDAR point clouds. 
\end{itemize}

\section{Related Work}
\label{sec:Related Work}
\subsection{LiDAR Perception}
LiDAR is widely used in various autonomous agents for collecting 3D point clouds from the environment. In the past, extensive research was dedicated to employing LiDAR for odometry~\cite{quenzel2021real} and mapping~\cite{droeschel2018efficient, zhong2023icra}. Given the inherent challenges of LiDAR, including data sparsity, noise, and outliers, researchers concentrated on developing filtering algorithms~\cite{rusu20113d} and robust point cloud registration~\cite{vizzo2023ral} to achieve accurate and efficient LiDAR-SLAM systems. With the advent of deep learning, researchers began focusing on the semantic properties of LiDAR data, with notable applications in object detection~\cite{wang2025adaptive} and semantic segmentation~\cite{li2025srkd}. Additionally, unlike dense representations such as images, the sparse nature of LiDAR point clouds presents unique challenges for models. To address these challenges, some researchers focus on estimating the gaps between sweeps and occluded regions from sparse point clouds. This has led to the development of semantic scene completion, an emerging technique in LiDAR perception.

\subsection{Semantic Scene Completion (SSC)}
Semantic scene completion (SSC) aims to jointly infer complex geometric structures and diverse semantic categories of a scene from partial observations. Since its introduction, various input data modalities, such as occupancy grids~\cite{roldao2020lmscnet}, images~\cite{cao2022monoscene}, and LiDAR-camera fusion~\cite{cao2024slcf}, have been explored. In parallel, a wide array of methodologies, including transformers~\cite{cao2025swasop}, bird's-eye view (BEV) assistance~\cite{cheng2021s3cnet}, and object-centric modeling~\cite{cao2025ocsop}, have been employed to advance the state of the art in this domain. However, these approaches generally operate on voxelized grids, which poses specific challenges for LiDAR point clouds, as voxelization can introduce quantization errors, leading to resolution loss and increased memory usage. In this work, we operate directly on point clouds, offering a more efficient and resolution-preserving method for handling LiDAR data.

\subsection{Denoising Diffusion Probabilistic Models}
Although diffusion models were originally discovered and proposed in the field of physics, DDPMs~\cite{ho2020denoising} were the first to apply this method to generative models. In subsequent research, Rombach et al.~\cite{rombach2022high} introduced latent diffusion models, where the diffusion process is performed in the latent space of the image. This significantly improved computational efficiency and reduced resource consumption, enabling the generation of high-quality and high-resolution images, marking a breakthrough in the field of artistic creation. Beyond artistic applications, diffusion models have been extended to spatiotemporal prediction tasks\cite{zeng2025enhancing} and LiDAR perception~\cite{nakashima2023lidar, zyrianov2022learning}, where 3D data is often projected onto range images, allowing methods developed for image domains to be directly applied. Notably, due to the higher demands for accuracy in robotics, controlling the generative process to achieve realistic results remains a significant challenge when applying diffusion models in this field. The recent LiDiff~\cite{nunes2024cvpr} directly applies diffusion models to 3D point clouds for scene completion. However, it still lacks the capability to model and process semantics simultaneously. In this work, we apply DDPM to semantic scene completion, to generate dense and accurate semantic scenes.

\section{Methodology}
\label{sec:Methodology}
\begin{figure*}[!ht]
	\vspace{.5em}
	\centering
	\includegraphics[width=\textwidth]{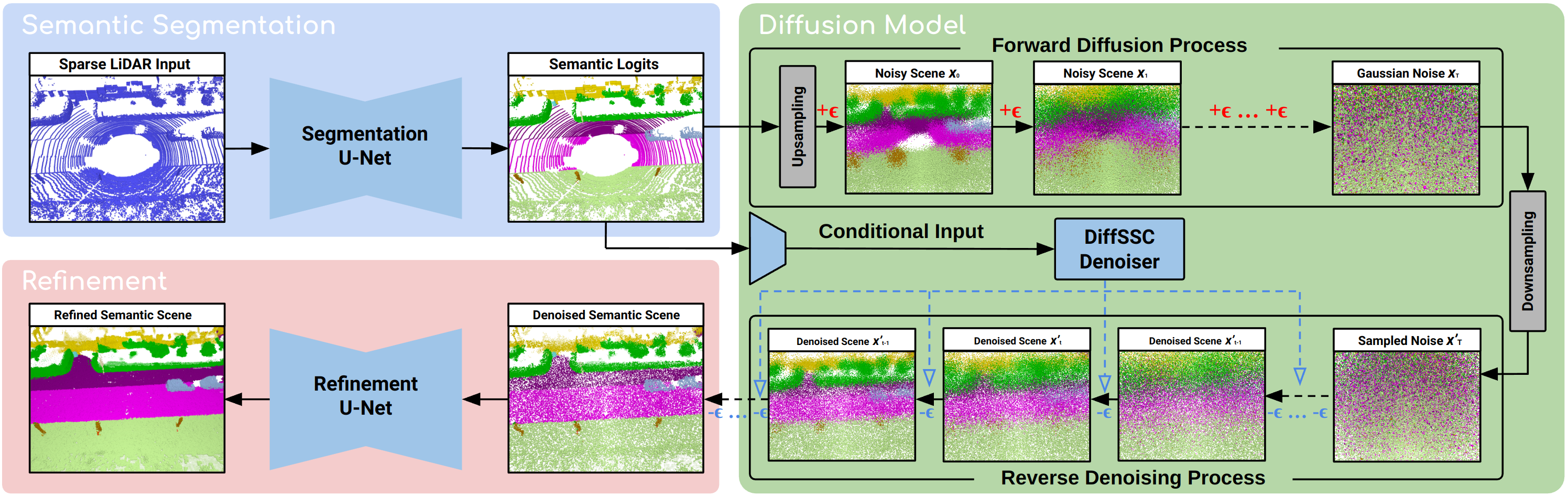}\vspace*{-.5em}
	\caption{The overall pipeline of DiffSSC. The raw LiDAR point cloud is semantically segmented using Cylinder3D~\cite{zhu2020cylindrical} to generate initial semantic logits. The semantic point cloud is then upsampled. These duplicated points undergo forward diffusion and reverse denoising. The original semantic point cloud serves as a conditional input, guiding the scene generation. To further enhance the generated scene, we introduce a refinement model based on MinkUNet~\cite{choy20194d, choy2019fully, choy2020high, gwak2020gsdn}, which increases the density of the point cloud.}
	\label{fig:diffssc}
	\vspace*{-.5em}
\end{figure*}

Given a raw LiDAR point cloud, our objective is to estimate a more complete semantic point cloud, including unobserved points with associated semantic labels within gaps and occluded regions. As illustrated in the Fig.~\ref{fig:diffssc}, we build a diffusion model supported by a semantic segmentation module and a refinement module. First, the raw LiDAR point cloud is semantically segmented using a Cylinder3D~\cite{zhu2020cylindrical} to generate initial semantic logits. Next, we upsample the semantic point cloud to increase point density for the diffusion process. The duplicated semantic points undergo a forward diffusion and a reverse denoising process to adjust their positions and semantics. Notably, the semantic point cloud also serves as a conditional input for the diffusion model, guiding the generation process. The generated scene includes semantic points located in gaps and occluded areas. To further enhance the quality of the generated scene, we designed a refinement model based on MinkUNet~\cite{choy20194d, choy2019fully, choy2020high, gwak2020gsdn} to densify the point cloud.

\subsection{Denoising Diffusion Probabilistic Models (DDPMs)}
Ho et al.~\cite{ho2020denoising} introduced DDPMs to produce high-quality images through iterative denoising from Gaussian noise. This promising capability is driven by a residual learning mechanism that efficiently captures the data distribution. Specifically, the process begins with a forward diffusion step, during which noise is gradually injected into the target data over $T$ steps. The model is then trained to estimate the noise injected at each step. By predicting and removing noise at time step $t$, the model generates results that closely approximate the raw data distribution.

\subsubsection{Forward Diffusion Process}
Assuming a sample $ \boldsymbol{x}_0 \sim q(\boldsymbol{x}) $ from a target data distribution, the diffusion process gradually adds noise to $ \boldsymbol{x}_0 $ over $ T $ steps, producing a sequence $ \boldsymbol{x}_1, \ldots, \boldsymbol{x}_T $. When $ T $ is large enough, $ q(\boldsymbol{x}_T)$ is approximately equal to a normal distribution $ \mathcal{N}(\boldsymbol{0}, \boldsymbol{I}) $. The intensity of noise added at each step is determined by the noise intensity factors $ \beta_1, \ldots, \beta_T $, which significantly influences the performance of the diffusion model. Specifically, at step $ t $, Gaussian noise amplified by $ \beta_t $ is sampled and added to $ \boldsymbol{x}_{t-1} $. In~\cite{ho2020denoising}, the noise parameter $\beta_t$ is determined using a linear schedule, starting from an initial value $\beta_0$ and linearly increasing over $T$ steps to a final value $\beta_T$. Subsequently, several improved noise schedules have been proposed, such as the cosine schedule~\cite{nichol2021improved} and the sigmoid schedule~\cite{kingma2021variational}. Due to the inefficiency of adding noise step by step, especially during batch loading, where the noise from different steps can be shuffled, one can simplify this process by sampling $ \boldsymbol{x}_t $ from $ \boldsymbol{x}_0 $ without computing the intermediate steps $ \boldsymbol{x}_1, \ldots, \boldsymbol{x}_{t-1}$. To achieve this, Ho et al.~\cite{ho2020denoising} define $ \alpha_t = 1 - \beta_t $ and $ \bar{\alpha}_t = \prod_{i=1}^{t} \alpha_i $, allowing $ \boldsymbol{x}_t $ to be sampled as:
\begin{align}
\boldsymbol{x}_t = \sqrt{\bar{\alpha}_t}\boldsymbol{x}_0 + \sqrt{1 - \bar{\alpha}_t} \boldsymbol{\epsilon}
\label{eqn:diffusion}
\end{align}
where $ \boldsymbol{\epsilon} \sim \mathcal{N}(\boldsymbol{0}, \boldsymbol{I}) $. It is important to note that when $ T $ is large enough, $ q(\boldsymbol{x}_T) $ approaches $ \mathcal{N}(\boldsymbol{0}, \boldsymbol{I}) $ because $ \bar{\alpha}_T $ tends to zero.

\subsubsection{Reverse Denoising Process}
The denoising process reverses diffusion and aims to recover the original sample $\boldsymbol{x}_0$ from Gaussian noise. This is accomplished by a denoiser, which estimates and removes the noise at each step. The reverse diffusion step can be formulated as:
\begin{equation*}
\boldsymbol{x}_{t-1} = \frac{1}{\sqrt{\alpha_t}} \left( \boldsymbol{x}_t - \frac{1-\alpha_t}{\sqrt{1-\bar{\alpha}_t}}\boldsymbol{\epsilon}_\theta(\boldsymbol{x}_t, t) \right) + \sigma_t \boldsymbol{\epsilon},
\end{equation*}
\begin{equation}
\text{with } \sigma_t^2 = \frac{1 - \bar{\alpha}_{t-1}}{1 - \bar{\alpha}_t} \beta_t
\label{eqn:denoising}
\end{equation}
where $\boldsymbol{\epsilon}_\theta(\boldsymbol{x}_t, t)$ is the noise estimated from $\boldsymbol{x}_t$ at step $t$. The process of generating the original data can be formulated as a Markov process that repeatedly calls the denoiser until $t=0$. At this point, the model generates a result that approximates $\boldsymbol{x}_0$. Due to the denoiser effectively learning the high quality of the data distribution $q(\boldsymbol{x}_T)$, the generated samples are of similarly high quality.

While the denoising process generates samples with quality similar to the dataset, it only produces random samples. Hence, the denoising process cannot control the generation of specific desired data, which poses challenges for certain downstream applications.~\cite{nichol2021improved} addresses this issue by introducing conditional inputs to guide the generation process. This advancement allows us to apply diffusion models to tasks like SSC.

\subsection{Diffusion Semantic Scene Completion}
Regarding the principles of DDPMs, we introduce its application in SSC. To focus on the main components, we assume that primary semantic segmentation has been obtained using Cylinder3D. In the context of the diffusion model, the input is a partial semantic point cloud $ \mathcal{X} = \{\boldsymbol{x}^1, \dots, \boldsymbol{x}^N\} $, where each semantic point $ \boldsymbol{x}^n $ is a tuple of a point position and a semantic probability vector $ (\boldsymbol{p}^n, \boldsymbol{s}^n) $. Here, $ \boldsymbol{p}^n \in \mathbb{R}^3 $ represents the 3D coordinates, and $ \boldsymbol{s}^n \in \Delta^{C-1} = \{\boldsymbol{s} \in \mathbb{R}^C \mid \sum_{i=1}^C s^i = 1, s^i \geq 0\} $ lies in the standard $ (C-1) $-dimensional simplex, assuming there are $ C $ classes in total. The output is the estimated complete point cloud $ \hat{\mathcal{Y}} = \{\hat{\boldsymbol{y}}^1, \dots, \hat{\boldsymbol{y}}^M\} $. We generate the reference $ \mathcal{Y} = \{\boldsymbol{y}^1, \dots, \boldsymbol{y}^M\} $ by fusing multiple frames with ground-truth semantic labels and then taking the corresponding region as the input scan $\mathcal{X}$. Our goal is to make the estimated $ \hat{\mathcal{Y}}$ as close as possible to the ground truth $ \mathcal{Y}$.

As mentioned in Sec.~\ref{sec:Introduction}, by learning scene priors, the model gains the ability to estimate a complete scene (posterior) from partial observations (likelihood). The diffusion model efficiently learns the distribution of the ground truth data, acquiring knowledge of the scene prior. To achieve this, we gradually inject noise into the ground truth $\mathcal{Y}$, resulting in $\mathcal{Y}_1, \dots, \mathcal{Y}_T$, until $\mathcal{Y}_T$ approximates a Gaussian distribution. However, the noise injection process in Eq.~\ref{eqn:diffusion} assumes that $\boldsymbol{x}_0$ is approximately isotropic, which is not suitable for LiDAR-scanned 3D scenes due to significant scale variations across the three spatial dimensions. While normalization can compress the scene into a more isotropic form, it also leads to a significant loss of fine details. To adapt noise injection for LiDAR-scanned 3D scenes,~\cite{nunes2024cvpr} apply local noise offsets at each point. This ensures that the noise intensity remains consistent across all spatial locations. Similarly, semantic categories in autonomous driving follow a long-tailed distribution~\cite{behley2019iccv}, indicating that their distribution exhibits anisotropy. Inspired by~\cite{nunes2024cvpr}, we adopt a local noise offset strategy for semantic noise injection. Specifically, we first scale the one-hot semantic encoding of the ground truth $\mathcal{Y}$ into the logit domain, then add noise offsets independently to each category's semantic logit, and subsequently restore the probabilistic semantic distribution via softmax. Combining the spatial and semantic domains, we propose a local anisotropic noise injection mechanism.
\begin{align}
\boldsymbol{y}_t^m = \boldsymbol{y}^m + \sqrt{1 - \bar{\alpha}_t} \boldsymbol{W} \boldsymbol{\epsilon}, \boldsymbol{W} =
\begin{bmatrix}
\sigma_{p}\boldsymbol{I}_3 & \boldsymbol{0} \\
\boldsymbol{0} & \sigma_{s}\boldsymbol{I}_C
\end{bmatrix}
\label{eqn:local diffusion}
\end{align}

Here, we employ an anisotropic scaling matrix $\boldsymbol{W}$, controlled by the scaling factors $\sigma_p$ and $\sigma_s$, to modulate the standard Gaussian noise $\boldsymbol{\epsilon}$, ensuring that it appropriately adapts to the scale differences in both the spatial and semantic logits domains. The modulated noise is then injected into the local semantic point $\boldsymbol{y}^m \in \mathbb{R}^{3+C}$. Given a specific time step $t \in [0,T]$, applying Eq.~\ref{eqn:local diffusion} to all points allows for computing the entire noised scene $\mathcal{Y}_t$ in a single step, eliminating the need for intermediate computations. This significantly reduces both memory consumption and computational time.

To enable the model to generate a corresponding complete semantic scene based on the current partial input, we encode the partial semantic point cloud $\mathcal{X}$ as a conditional input, which is then fed into the model to guide the point cloud generation process. Thus, the denoiser integrates the noised scene $\mathcal{Y}_t$, the conditional input $\mathcal{X}$, and the time step $t$, which indicates the intensity of noise, to estimate the noise $\boldsymbol{\epsilon}_{\theta}(\mathcal{Y}_t, \mathcal{X}, t)$.

Based on our residual learning mechanism, we employ the $L_2$ loss to regularize the local discrepancy between the estimated noise and the real noise, rather than comparing the generated scene and the target scene.
\begin{align}
L_2 = \left\lVert \sqrt{1 - \bar{\alpha}_t} \boldsymbol{W} \boldsymbol{\epsilon} - \boldsymbol{\epsilon}_{\theta}(\mathcal{Y}_t, \mathcal{X}, t) \right\rVert^2
\label{eqn:local loss}
\end{align}
Therefore, to generate the final scene, we additionally remove the estimated noise from the sample, followed by transforming the semantic logits into a semantic probability vector via the softmax function.

Besides the local loss $L_2$ commonly used in DDPM models, ~\cite{nunes2024cvpr} propose a global regularization for the mean and variance of the estimated noise, enforcing it to follow the statistical properties of the injected noise, i.e., a Gaussian distribution. While first- and second-order moments effectively regularize noise in the spatial domain, the estimated noise in the semantic domain tends to exhibit a skewed distribution due to the long-tailed nature of semantic data in autonomous driving. Therefore, we introduce skewness regularization as the third-order constraint to improve the noise distribution in the semantic domain. Thus, the overall loss is formulated as follows:
\begin{equation*}
L = L_2 + \lambda_p (L_{\text{p,mean}} + L_{\text{p,var}}) + \lambda_s (L_{\text{s,mean}} + L_{\text{s,var}} + L_{\text{s,skew}})
\end{equation*}
\begin{equation*}
L_{\text{p,mean}} = \bar{\boldsymbol{\epsilon}_p}^2, \quad 
L_{\text{p,var}} = (\hat{\boldsymbol{\epsilon}_p} -1)^2
\end{equation*}
\begin{equation}
L_{\text{s,mean}} = \bar{\boldsymbol{\epsilon}_s}^2, \quad 
L_{\text{s,var}} = (\hat{\boldsymbol{\epsilon}_s} -1)^2, \quad 
L_{\text{s,skew}} = \tilde{\boldsymbol{\epsilon}_s}^2
\label{eqn:regularization}
\end{equation}
where $\boldsymbol{\epsilon}_p$ and $\boldsymbol{\epsilon}_s$ correspond to the spatial and semantic domains, respectively, with their contributions weighted by $\lambda_p$ and $\lambda_s$. Meanwhile,
$\bar{\boldsymbol{\epsilon}_{\theta}}$, $\hat{\boldsymbol{\epsilon}_{\theta}}$, and $\tilde{\boldsymbol{\epsilon}_{\theta}}$ represent the mean, standard deviation, and skewness of the estimated noise $\boldsymbol{\epsilon}_{\theta}$, respectively. Compared to the local term $L_2$, the global term is used to regularize the statistical properties of the noise, ensuring that it approximates a Gaussian distribution.

\subsection{Denoiser Architecture}
\begin{figure}[tb]
    \centering
    \includegraphics[width=0.45\textwidth]{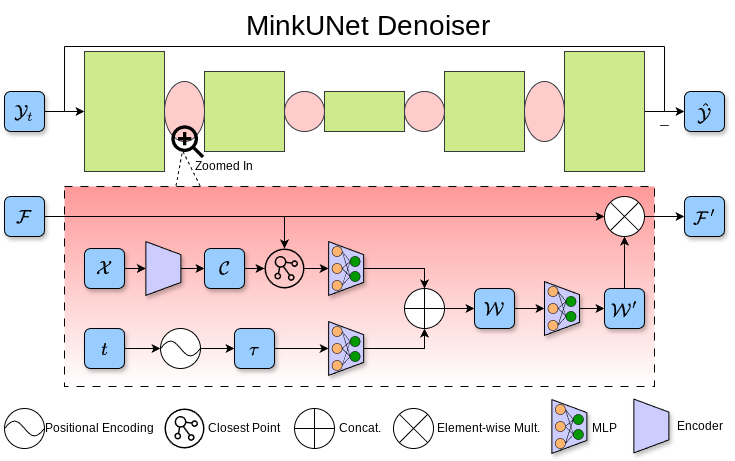}
    \caption{Architecture of our MinkUNet Denoiser. As shown in the red area, we design information fusion layers and insert them between MinkUNet blocks to integrate the conditional input and step information, guiding the generation of the point cloud.}
    \label{fig:denoiser}
    \vspace{-1.em}
\end{figure}
As shown in Fig.~\ref{fig:denoiser}, the denoiser is based on the MinkUNet architecture~\cite{choy20194d, choy2019fully, choy2020high, gwak2020gsdn}. Given the feature $\mathcal{F}$ extracted from a layer of MinkUNet, we integrate the conditional input and step information between layers to obtain the fused feature $\mathcal{F}'$. The raw semantic point cloud $\mathcal{X}$ is encoded as a conditional input $\mathcal{C}$. To embed the most relevant conditional input into the feature space, a closest point algorithm is employed to effectively align the conditional input with the features. Simultaneously, the step $t$ is encoded as $\tau$ using sinusoidal positional encodings. After passing through an MLP individually, the conditional input and step information are concatenated to form the weight $\mathcal{W}$. To align the dimensions with the feature $\mathcal{F}$, $\mathcal{W}$ is processed through an MLP to produce $\mathcal{W}'$. Finally, $\mathcal{W}'$ and $\mathcal{F}$ are element-wise multiplied to form the refined feature $\mathcal{F}'$, which is then passed to the next layer.

\subsection{Refinement} 
Inspired by Lyu et al.~\cite{lyu2021conditional}, we design a refinement and upsampling scheme based on MinkUNet to further enhance the density of the diffusion model's output. This module predicts $k$ bias $b_k \in \mathbb{R}^3 $ for each point position in the completed scene, while the semantics are propagated to the biased points. The refinement module offers a marginal improvement in scene quality, but it functions as interpolating points in the gaps, rather than learning to predict missing geometry and semantics. The main contribution is made by the diffusion model, as will be demonstrated in the ablation study.

\section{Experiments}
\label{sec:Experiment}
\definecolor{carColor}{RGB}{100,150,245}
\definecolor{bicycleColor}{RGB}{100,230,245}
\definecolor{motorcycleColor}{RGB}{30,60,150}
\definecolor{truckColor}{RGB}{80,30,180}
\definecolor{othervehicleColor}{RGB}{0,0,255}
\definecolor{personColor}{RGB}{255,30,30}
\definecolor{bicyclistColor}{RGB}{255,40,200}
\definecolor{motorcyclistColor}{RGB}{150,30,90}
\definecolor{roadColor}{RGB}{255,0,255}
\definecolor{parkingColor}{RGB}{255,150,255}
\definecolor{sidewalkColor}{RGB}{75,0,75}
\definecolor{othergroundColor}{RGB}{175,0,75}
\definecolor{buildingColor}{RGB}{255,200,0}
\definecolor{fenceColor}{RGB}{255,120,50}
\definecolor{vegetationColor}{RGB}{0,175,0}
\definecolor{trunkColor}{RGB}{135,60,0}
\definecolor{terrainColor}{RGB}{150,240,80}
\definecolor{poleColor}{RGB}{255,240,150}
\definecolor{trafficsignColor}{RGB}{255,0,0}

\subsection{Benchmark Result}
To conduct a comprehensive comparison with other leading SSC methods, we first evaluate our model on the SemanticKITTI~\cite{behley2019iccv} Benchmark. SemanticKITTI is a widely used autonomous driving dataset that provides point-wise annotations on raw LiDAR point clouds, extending the original KITTI dataset for semantic understanding tasks. The SSC Benchmark is a subtask within SemanticKITTI, focusing on predicting both the semantic class and occupancy status of each voxel within a grid volume. To obtain the ground truth, the annotated sequential scans are first accumulated and then chopped based on a predefined range represented in the LiDAR sensor's coordinate system: $V_{\text{kitti}} = \{(x, y, z) \mid x \in [0, 51.2] \text{ m}, y \in [-25.6, +25.6] \text{ m}, z \in [-3.2, +3.2] \text{ m}\}$. The extracted region is then voxelized into a $ 256 \times 256 \times 32 $ grid volume, where each voxel represents a $ 0.2^3 \, \text{m}^3 $ cube in the real world. Although our method operates directly on point clouds, the discrete nature of point clouds makes it challenging to directly evaluate performance using traditional IoU metrics, which are designed for continuous spatial regions. To address this, we voxelized our results and employed IoU for scene completion and mIoU for semantic scene completion evaluation. While voxelization introduces quantization errors that may slightly degrade our model's performance, this approach ensures a fair and meaningful comparison.

\begin{table}[h]
\caption{Quantitative results on the SemanticKITTI benchmark.}
\label{tab:benchmark}
\centering
\setlength{\tabcolsep}{6pt}
\begin{threeparttable}
\small
\begin{tabular}{c|c|cc}
  \toprule
  Method & Reference & IoU($\%$) & mIoU($\%$)\\
  \midrule
  LMSCNet~\cite{roldao2020lmscnet} & 3DV'20 & 55.3 & 17.0 \\
  S3CNet~\cite{cheng2021s3cnet} & CoRL'21 & 45.0 & 29.5 \\
  SSA-SC~\cite{yang2021semantic} & IROS'21 & 58.8 & 23.5\\
  JS3C-Net~\cite{yan2021sparse}& AAAI'21 & 56.6 & 23.8\\
  LODE~\cite{li2023lode}& ICRA'23 & 51.2 & 23.4 \\
  SCPNet~\cite{xia2023scpnet} & CVPR'23 & 56.1 & \underline{36.7}\\
  TALoS~\cite{jang2024talos} & NeurIPS'24 & \underline{60.2} & \textbf{37.9} \\
  \midrule
  Ours & - & \textbf{63.4} & 27.4 \\
  \bottomrule
\end{tabular}
\end{threeparttable}
\vspace*{1ex} 

\footnotesize{IoU is used to evaluate only the occupancy status, while mIoU assesses performance across all semantic classes. \textbf{Best} and \underline{second best} results are highlighted.}
\vspace*{-0.5em}
\end{table}

We follow the official dataset split: sequences 00-07 and 09-10 are used for training, 08 for validation, and 11-21 for testing. Model performance is evaluated through the official online benchmark server. The model is trained on an NVIDIA A6000 GPU for 20 epochs. For the diffusion parameters, we employ a cosine schedule to modulate the intensity of noise at each step. Specifically, we set $\beta_0 = 3.5 \times 10^{-5}$ and $\beta_T = 0.007$, with the number of diffusion steps $T = 1000$, and define $\beta_1, \dots, \beta_{T-1}$ using the following equation.
\begin{align}
\beta_t = \beta_0 + \frac{1}{2} \left( 1 + \cos\left(\frac{t}{T} \cdot \pi \right)\right) \cdot (\beta_T - \beta_0)
\label{eqn:cosine}
\end{align}
We set the ratio of global regularization to $\lambda_p = 5.0$ and $\lambda_s = 4.0$. Additionally, we define the scaling factors as $\sigma_p = 1.0$ and $\sigma_s = 0.2$. As shown in Tab.~\ref{tab:benchmark}, our method outperforms all state-of-the-art LiDAR-based approaches in scene completion, indicating that the model effectively captures geometric information. However, for semantic scene completion (measured by mIoU), a obvious gap remains compared to SCPNet and TALoS. It is important to note that both methods leverage additional information beyond single-frame input. Specifically, SCPNet utilizes multi-frame knowledge distillation to enhance the prediction of the current frame, while TALoS employs a test-time adaptation strategy, incorporating multi-frame observations and future data to refine its outputs. In contrast, DiffSSC relies solely on single-frame information and does not incorporate online model adjustments.

While the SemanticKITTI SSC benchmark provides a valuable framework for evaluating SSC models, its design is notably influenced by early indoor SSC tasks. These early methods mainly relied on RGB-D cameras, which inherently limited perception to front-facing scenes. To leverage the technical foundations of these methods and simplify their transition to outdoor environments, SemanticKITTI restricts the LiDAR field of view to the front half, covering only the range $[-90^\circ, +90^\circ]$ in the sensor coordinate system, while completely ignoring the rear part of the scene. Although this simplification facilitated early exploration of outdoor SSC, it fundamentally deviates from the natural properties of LiDAR data and the core requirements of autonomous driving tasks. Rear-view perception is equally critical for driving safety, especially for tasks like lane changes, reversing, and obstacle avoidance. Since LiDAR sensors capture data through continuous $360^\circ$ rotations without directional bias, restricting input to the front half disrupts the spatial consistency of the data and limits the model's ability to fully understand the driving environment. To address this limitation, we extend the scene completion task to a full $360^\circ$ panoramic view, preserving the intrinsic characteristics of LiDAR data and providing a more comprehensive representation of real-world autonomous driving scenarios.

\subsection{Extended Experiment on Panoramic Scenarios}
\subsubsection{Panoramic Settings}
In the raw SSC setting of SemanticKITTI, the scene is limited to a cuboid region $V_{\text{kitti}}$, covering the range $[-90^\circ, +90^\circ]$. To extend this to the full $[-180^\circ, +180^\circ]$ panoramic range while preserving spatial symmetry, the panoramic volume is defined as the combination of two SemanticKITTI volumes facing forward and backward. Specifically, in the LiDAR's local coordinate system, this is formulated as: $V_{\text{pano}} = \{(x, y, z) \mid x \in [-51.2, 51.2] \text{ m}, y \in [-25.6, +25.6] \text{ m}, z \in [-3.2, +3.2] \text{ m}\}$. The front and rear halves of the panoramic volume are designed to be identical to the original SemanticKITTI volume. Given the similar statistical characteristics of LiDAR data in the front and rear regions, retaining the original SemanticKITTI volume settings allows baseline methods to be seamlessly transferred to the panoramic setting without significant performance degradation.

We generate the ground truth following the guidelines of SemanticKITTI. First, using the pose information of each frame, we construct a global map by aggregating the semantic LiDAR sweeps within the sequence. Next, we extract the region within $V_{\text{pano}}$, with the LiDAR positioned at its center. Additionally, unknown areas defined by the raw dataset are mapped into $V_{\text{pano}}$ using the known poses, and these regions are excluded from the evaluation.

Besides SemanticKITTI, we also conduct panoramic experiments on the SSCBench-KITTI360 dataset~\cite{li2023sscbench}. SSCBench-KITTI360 is another SSC benchmark derived from KITTI-360~\cite{liao2022kitti}, with semantic information aligned to the SemanticKITTI format. This consistency allows SSC methods evaluated on SemanticKITTI to be seamlessly transferred to the KITTI-360 scenario. However, both benchmarks only use the front half of the LiDAR scan as input. To address this, we apply the same panoramic processing approach to SSCBench-KITTI360 as we did for SemanticKITTI.

\subsubsection{Baselines}
We compare our approach against LMSCNet~\cite{roldao2020lmscnet}, JS3C-Net~\cite{yan2021sparse}, and LODE~\cite{li2023lode}. Both LMSCNet and JS3C-Net take the front half of the quantized LiDAR sweep as input and are evaluated on the SSC benchmark of SemanticKITTI. LODE primarily focuses on geometry completion using implicit representations; however, to demonstrate its flexibility, the authors also report results with extended semantic parsing.

To enable these baselines to predict panoramic scenes, we split the $360^\circ$ LiDAR point cloud into two halves and feed them separately into the baseline models. The front half follows the same settings as in SemanticKITTI, while the rear half is rotated by $180^\circ$ before being passed to the models. After obtaining predictions for both halves, we concatenate them to form the complete panoramic scene. Although the baselines were trained solely on the front part of the scene, the statistical characteristics of LiDAR data in the front and rear regions are similar. This suggests that models trained on the front half remain effective when applied to the rear region. This approach minimizes performance degradation due to domain shift, ensuring a fair comparison. Additionally, we directly utilized the official code and pretrained checkpoints from the baselines to predict the panoramic scenes, further maintaining consistency in evaluation. 

While these baselines have reported results on SemanticKITTI, they had not previously been tested on SSCBench-KITTI360~\cite{li2023sscbench}. To supplement our evaluation, we ran these baselines on SSCBench-KITTI360 without fine-tuning. Since the semantic labels and the overall pipeline in SSCBench-KITTI360 are consistent with SemanticKITTI, the baselines could be seamlessly applied to this dataset.

\subsubsection{Training and Inference}
Since our method operates directly on point clouds rather than voxel-based volumes, the processing of training pairs differs from that of the baselines. Although we generate the ground truth by aggregating the semantic LiDAR sweeps and extracting point clouds within the $V_{\text{pano}}$, we do not further voxelize the data. To facilitate better diffusion and learning, we define the generated scene range during both training and inference as a spherical area centered on the LiDAR with a radius of 60 meters, i.e., $V_{\text{sphere}} = \{(x, y, z) \mid \sqrt{x^2 + y^2 + z^2} \leq 60 \, \text{m} \}$.
During evaluation, we voxelize the predicted point cloud scene and extract the portion within $V_{\text{pano}}$. Thus, while the generated scene range differs from $V_{\text{pano}}$, the evaluation region is restricted to $V_{\text{pano}}$, ensuring consistency with the baselines. Moreover, the model does not leverage any information outside $V_{\text{pano}}$ but within $V_{\text{sphere}}$ during the learning process, ensuring a fair comparison with the baselines. Our model is trained and validated purely on SemanticKITTI, using sequences 00-07 for training and sequences 09-10 for validation. We evaluate our model on the official validation sets of both datasets: sequence 08 of SemanticKITTI and sequence 07 of SSCBench-KITTI360. Since the baselines were not fine-tuned on SSCBench-KITTI360, we likewise did not perform any fine-tuning on SSCBench-KITTI360.

\subsubsection{Experimental Results}
Based on the experimental settings described above, we compare the performance of existing SSC methods with our approach in Tab.~\ref{tab:quantitative}. Although voxel-based evaluation introduces quantization errors, the output of our diffusion model surpasses all baselines, demonstrating the effectiveness of our method.

\begin{table}[h]
\caption{Quantitative results of the panoramic experiments on the SemanticKITTI and SSCBench-KITTI360 validation sets.}
\label{tab:quantitative}
\centering
\renewcommand{\arraystretch}{1.2} 
\begin{threeparttable}
\small
\begin{tabular}{c|cc|cc}
  \toprule
  \multirow{2}{*}{Method} & \multicolumn{2}{c|}{SemanticKITTI} & \multicolumn{2}{c}{SSCBench-KITTI360} \\
  & IoU($\%$) & mIoU($\%$) & IoU($\%$) & mIoU($\%$) \\
  \midrule
  LMSCNet~\cite{roldao2020lmscnet} & 48.2 & 15.4 & 33.6 & 13.5\\
  JS3C-Net~\cite{yan2021sparse} & 51.3 & 21.4 & 35.6 & 17.0\\
  LODE~\cite{li2023lode} & 50.6 & 18.2 & 38.2 & 15.4\\
  \midrule
  Ours &\textbf{60.3} & \textbf{26.7} & \textbf{47.3} & \textbf{20.4}\\
  \bottomrule
\end{tabular}
\end{threeparttable}
\vspace*{1ex} 

  \footnotesize{\textbf{Best} results are highlighted in bold.}
\end{table}

\begin{figure*}[h]
  \centering
  \begin{tikzpicture}
    \node[anchor=south west,inner sep=0] (image) at (0,0) {\includegraphics[width=\textwidth]{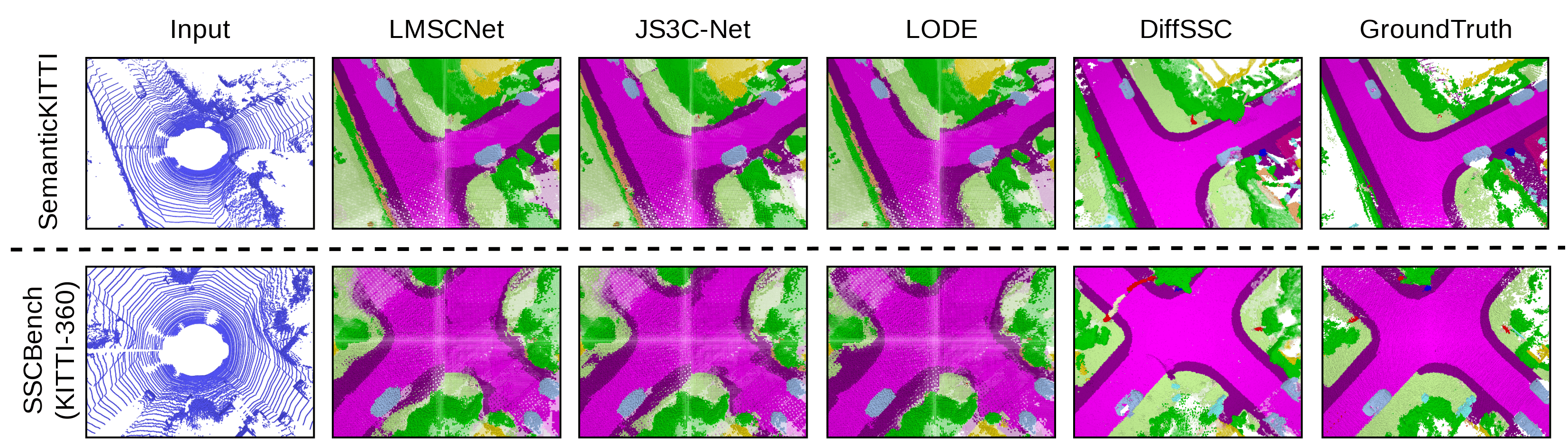}};
    \def\legendWidth{2.0}
    \def\legendHeight{0.2}
    \def\legendSpacing{0.15}
    \def\yOffset{-0.3}
    
   \draw[fill=carColor] ($ (image.south west) + (0.2, \yOffset)$) rectangle ++(0.2, 0.2) node[right, yshift=-0.1cm] {\small car};
	\draw[fill=bicycleColor] ($ (image.south west) + (1.2, \yOffset)$) rectangle ++(0.2, 0.2) node[right, yshift=-0.1cm] {\small bicycle};
    \draw[fill=motorcycleColor] ($ (image.south west) + (2.7, \yOffset)$) rectangle ++(0.2, 0.2) node[right, yshift=-0.1cm] {\small motorcycle};
    \draw[fill=truckColor] ($ (image.south west) + (4.7, \yOffset)$) rectangle ++(0.2, 0.2) node[right, yshift=-0.1cm] {\small truck};
    \draw[fill=othervehicleColor] ($ (image.south west) + (5.9, \yOffset)$) rectangle ++(0.2, 0.2) node[right, yshift=-0.1cm] {\small other-vehicle};
    \draw[fill=personColor] ($ (image.south west) + (8.1, \yOffset)$) rectangle ++(0.2, 0.2) node[right, yshift=-0.1cm] {\small person};
    \draw[fill=bicyclistColor] ($ (image.south west) + (9.5, \yOffset)$) rectangle ++(0.2, 0.2) node[right, yshift=-0.1cm] {\small bicyclist};
    \draw[fill=motorcyclistColor] ($ (image.south west) + (11.3, \yOffset)$) rectangle ++(0.2, 0.2) node[right, yshift=-0.1cm] {\small motorcyclist};
    \draw[fill=roadColor] ($ (image.south west) + (13.4, \yOffset)$) rectangle ++(0.2, 0.2) node[right, yshift=-0.1cm] {\small road};
    \draw[fill=parkingColor] ($ (image.south west) + (14.5, \yOffset)$) rectangle ++(0.2, 0.2) node[right, yshift=-0.1cm] {\small parking};
    \draw[fill=sidewalkColor] ($ (image.south west) + (16.0, \yOffset)$) rectangle ++(0.2, 0.2) node[right, yshift=-0.1cm] {\small sidewalk};
    
    \draw[fill=othergroundColor] ($ (image.south west) + (2.7, \yOffset-\legendHeight - \legendSpacing)$) rectangle ++(0.2, 0.2) node[right, yshift=-0.1cm] {\small other-ground};
    \draw[fill=buildingColor] ($ (image.south west) + (4.8, \yOffset-\legendHeight - \legendSpacing)$) rectangle ++(0.2, 0.2) node[right, yshift=-0.1cm] {\small building};
        \draw[fill=fenceColor] ($ (image.south west) + (6.3, \yOffset-\legendHeight - \legendSpacing)$) rectangle ++(0.2, 0.2) node[right, yshift=-0.1cm] {\small fence};
    \draw[fill=vegetationColor] ($ (image.south west) + (7.4, \yOffset-\legendHeight - \legendSpacing)$) rectangle ++(0.2, 0.2) node[right, yshift=-0.1cm] {\small vegetation};
    \draw[fill=trunkColor] ($ (image.south west) + (9.2, \yOffset-\legendHeight - \legendSpacing)$) rectangle ++(0.2, 0.2) node[right, yshift=-0.1cm] {\small trunk};
    \draw[fill=terrainColor] ($ (image.south west) + (10.4, \yOffset-\legendHeight - \legendSpacing)$) rectangle ++(0.2, 0.2) node[right, yshift=-0.1cm] {\small terrain};
    \draw[fill=poleColor] ($ (image.south west) + (11.7, \yOffset-\legendHeight - \legendSpacing)$) rectangle ++(0.2, 0.2) node[right, yshift=-0.1cm] {\small pole};
    \draw[fill=trafficsignColor] ($ (image.south west) + (12.8, \yOffset-\legendHeight - \legendSpacing)$) rectangle ++(0.2, 0.2) node[right, yshift=-0.1cm] {\small traffic-sign};
  \end{tikzpicture}
	
	\vspace*{-1ex}

  \caption{Qualitative results of the panoramic experiments on the SemanticKITTI and SSCBench-KITTI360 validation sets. All 19 classes are displayed without empty spaces. Predicted points located in unknown regions are visualized with $20\%$ opacity.}
  \label{fig:visualization}
\end{figure*}

Qualitative results are presented in Fig.~\ref{fig:visualization}. To highlight the advantages of our approach, which operates directly on point clouds, we visualize samples from both the SemanticKITTI and SSCBench-KITTI360 datasets in point cloud form. For voxel-based methods, point clouds are generated by sampling the center point of each occupied voxel. As shown in Fig.~\ref{fig:visualization}, our DiffSSC model predicts more accurate semantic segmentation of the background and offers a more precise representation of foreground shapes. Moreover, voxel-based baselines, which estimate the scene using two halves of a LiDAR sweep, exhibit discontinuities at the boundary between the front and rear parts. This further underscores the importance of learning from panoramic LiDAR data.


\subsection{Ablation Studies}

\begin{table*}[t]
\caption{The results of ablation studies for the proposed DiffSSC. The performance is evaluated on SemanticKITTI and SSCBench-KITTI360 validation sets. \textbf{Best} results are highlighted.}
\label{tab:ablation}
\centering
\setlength{\tabcolsep}{1.3pt}
\begin{threeparttable}
\small
\begin{tabular}{c|c|ccc|ccc|cc|cc|cc}
  \toprule
   & \multirow{2}{*}{Method} & \multicolumn{3}{c|}{Module} &\multicolumn{3}{c|}{Noise Schedule} & \multicolumn{2}{c|}{Regularization} & \multicolumn{2}{c|}{SemanticKITTI} & \multicolumn{2}{c}{SSCBench-KITTI360} \\
   & & Segmentation & Diffusion & Refinement & Linear & Sigmoid & Cosine & Local & Global & IoU($\%$) & mIoU($\%$) & IoU($\%$) & mIoU($\%$) \\
   \midrule
   Ours & A &$\checkmark$ &$\checkmark$ &$\checkmark$ & -  &  - & $\checkmark$ & $\checkmark$ & $\checkmark$ & \textbf{60.3} & \textbf{26.7} & \textbf{47.3} & \textbf{20.4}\\
  \midrule
  \multirow{2}{*}{Module-level} &B & $\checkmark$ & - & $\checkmark$ & - & - & - & - & - &23.4 &7.6 &20.7 &7.2\\
  &C & $\checkmark$ & $\checkmark$ & - & - & - & $\checkmark$ & $\checkmark$ & $\checkmark$ & 58.7 & 26.3 & 42.1 & 19.3\\
  \midrule
  \multirow{3}{*}{Policy-level}  &D & $\checkmark$&$\checkmark$ & - & $\checkmark$ & - & - & $\checkmark$ & $\checkmark$ & 52.6 & 20.7 & 37.1 & 16.3\\
  &E & $\checkmark$& $\checkmark$& - &  - & $\checkmark$ & - & $\checkmark$ & $\checkmark$ & 56.9 & 25.8 & 40.9 & 18.5\\
   &F &$\checkmark$ &$\checkmark$ & - & - & - & $\checkmark$ & $\checkmark$ & - & 40.3 &10.9 & 30.7 & 9.6\\
   
  \bottomrule
\end{tabular}
\end{threeparttable}
\vspace*{1ex} 
\vspace*{-1.0em}
\end{table*}

To systematically analyze the contribution of each component in our model, particularly the core diffusion model and key learning strategies, we ablate our method on SemanticKITTI and SSCBench-KITTI360, with results summarized in Tab.~\ref{tab:ablation}. Our model architecture comprises three primary modules: semantic segmentation, diffusion, and refinement. We performed module-level ablations to evaluate the individual contributions of each submodule. Additionally, we conducted policy-level ablations to examine the influence of two critical factors on the diffusion model: the noise schedule function and global regularization. 

\subsubsection{Module-level}
The semantic segmentation module is essential for providing initial semantic priors. Therefore, we focused on analyzing the individual contributions of the diffusion and refinement modules. In Method B, we removed the core diffusion module from our pipeline, directly refining the output of Cylinder3D. Without the diffusion step, the noise schedule and regularization become irrelevant. This resulted in poor performance, indicating that refinement alone cannot effectively predict unknown areas in the scene. In Method C, we used only the diffusion model without refinement. While performance slightly dropped compared to the full Method A, this suggests that refinement improves scene densification but relies on accurate predictions from the diffusion model.

\subsubsection{Policy-level}
As mentioned in Sec.~\ref{sec:Methodology}, the noise schedule determines the intensity of noise injected at each step, commonly including linear, cosine, and sigmoid schedules. We investigated the impact of different noise schedules on the diffusion process. To isolate this effect, we removed the refinement module, allowing a clearer view of how the noise schedule influences diffusion. Results are shown in Tab.~\ref{tab:ablation}. Comparing Method C, D, and E, we observe that the linear schedule, being the simplest form, performs significantly worse than the other two schedules. The cosine schedule, with its S-shaped curve and precise control over noise introduction, balances faster convergence with high final generation quality, achieving the best results. The sigmoid schedule also shows competitive performance, slightly lagging behind the cosine schedule. Therefore, we adopted the cosine schedule in our main results.


We also investigated the impact of global regularization on model performance. By setting $\lambda_p = 5.0$ and $\lambda_s = 4.0$, we removed global regularization in Method F. Compared to Method C ($\lambda_p = \lambda_s = 0$), the model exhibited poorer performance, highlighting the advantages of incorporating global regularization.

\section{Conclusions and Outlook}
\label{sec:Conclusion and Outlook}
We proposed DiffSSC, a novel SSC approach based on DDPMs. It takes raw LiDAR point clouds as input and jointly predicts missing points along with their semantic labels, thereby extending the application boundaries of diffusion models. We evaluated our method on two autonomous driving datasets, achieving state-of-the-art performance. In future work, we plan to explore integrating cross-modal signals and prompt-guided learning to enhance scene understanding~\cite{wu2025llm, wuimgfu, wu2025prompt}. We will also explore strategies for improving inference efficiency and reducing resource consumption, inspired by recent advances in lightweight modeling~\cite{li2024sglp, li2025frequency}.

\bibliographystyle{IEEEtran}
\bibliography{literature}



\end{document}